\pdfoutput=1
\documentclass[11pt]{article}

\usepackage[]{emnlp2021}

\usepackage{times}
\usepackage{latexsym}

\usepackage[T1]{fontenc}
\usepackage[utf8]{inputenc}
\usepackage{graphicx}

\usepackage{microtype}
\usepackage{multirow}
\usepackage{tablefootnote}
\usepackage[OT6,T1]{fontenc}
\usepackage[utf8]{inputenc}
\usepackage[english,french]{babel}

\newcommand{\armenian}{\fontencoding{OT6}\fontfamily{cmr}\selectfont}
\DeclareTextFontCommand{\textarmenian}{\armenian}
\newcounter{licntr}			% linguistic item counter

\newcommand{\stepli}{\refstepcounter{licntr}}	% increment
\newcommand{\lival}{\thelicntr}			% return value
\newenvironment{liout}{
		\stepli			% step li counter
		\samepage			% prevent page break
		\begin{list}{		% begin definition
			(\lival)\hfill}{} 	% left adjust number
			\samepage
			\item 			% dummy item
		}{ \end{list}}

\newenvironment{li*}{
	\stepli
	\samepage				% prevent page break
	\begin{trivlist} \item[] (\lival) \end{trivlist}
	\begin{trivlist} 
	{\samepage \item[]}
	}{\end{trivlist}}					% space

\newcounter{alphcntr}		% will be reset in each new lis environment

\newcommand{\bi}{\begin{itemize}}		 
\newcommand{\ei}{\end{itemize}}		 

\newcommand{\be}{\begin{enumerate}}		 
\newcommand{\ee}{\end{enumerate}}

\newcommand{\bc}{\begin{center}}		
\newcommand{\ec}{\end{center}}

\newlength{\astspace}				% initialise length command

\newcounter{figcntr}			% figiom counter

\newcommand{\stepfig}{\refstepcounter{figcntr}}	% increment
\newcommand{\figval}{\thefigcntr}		% return value
\newenvironment{figout}{
		\stepfig			% step fig counter
		\begin{list}{		% begin definition
			(\figval)\hfill}{} 	% left adjust number
			\item 			% dummy item
		}{ \end{list}}

\renewcommand{\thefigcntr}{\Alph{figcntr}}

\newcounter{axcntr}			% axiom counter

\newcommand{\stepax}{\refstepcounter{axcntr}}	% increment
\newcommand{\axval}{\theaxcntr}		% return value
\newenvironment{axout}{
		\stepax			% step ax counter
		\begin{list}{		% begin definition
			(\axval)\hfill}{} 	% left adjust number
			\item 			% dummy item
		}{ \end{list}}

\renewcommand{\theaxcntr}{\Roman{axcntr}}

\newcounter{alphacntr}		% will be reset in each new ax environment

\title{NeuTral Rewriter: A Rule-Based and Neural Approach to Automatic Rewriting into Gender-Neutral Alternatives}

\author{Eva Vanmassenhove \and Chris Emmery \and Dimitar Shterionov\\
  Department of Cognitive Science and Artificial Intelligence \\
  Tilburg University \\
  The Netherlands \\
  \texttt{e.o.j.vanmassenhove@tilburguniversity.edu}\\
  \texttt{c.d.emmery@tilburguniversity.edu}\\
  \texttt{d.shterionov@tilburguniversity.edu}
  }
  
\begin{document}
\maketitle
\begin{abstract}

% Recent years have seen an increasing need for gender-neutral and inclusive language. This need is reflected, among others, by a surge in the use of \textit{singular they}, currently endorsed as part of APA style as the generic and gender-neutral pronoun. Within the field of NLP, there are furthermore various monolingual and bilingual use cases where gender neutral and inclusive language is appropriate, if not preferred, due to ambiguity/uncertainty in terms of the gender of the referents. 

Recent years have seen an increasing need for gender-neutral and inclusive language. Within the field of NLP, there are various mono- and bilingual use cases where gender inclusive language is appropriate, if not preferred due to ambiguity or uncertainty in terms of the gender of referents. In this work, we present a rule-based and a neural approach to gender-neutral rewriting for English along with manually curated synthetic data (\textit{WinoBias+}) and natural data (OpenSubtitles and Reddit) benchmarks. A detailed manual and automatic evaluation highlights how our NeuTral Rewriter, trained on data generated by the rule-based approach, obtains word error rates (WER) below 0.18\% on synthetic, in-domain and out-domain test sets.

\end{abstract}

\section{Introduction}\label{sec:introduction}
Recent years have seen an increasing need for gender-neutral and inclusive language. This need is reflected, among others, by a surge in the use of \textit{singular they},\footnote{The pronoun `they' was announced word of the year in 2019 according to Merriam Webster \url{https://www.nytimes.com/2019/12/10/us/merriam-webster-they-word-year.html}} currently endorsed as part of APA style as the generic and gender-neutral pronoun.~\footnote{\url{https://apastyle.apa.org/}} Within the field of Natural Language Processing (NLP), there are various monolingual and bilingual use cases where gender neutral and inclusive language is appropriate, if not preferred due to e.g. ambiguity in terms of the gender of referents. Section \ref{sec:usecases} provides a short outline of potential NLP use cases.

% Guidelines on gender-neutral and inclusive language have also been published by major (supra)national organizations and institutions (e.g. European Parliament, United Nations). Given that Natural Language Processing (NLP) technology

To support these use cases, we present a rule-based and a neural approach to gender-neutral rewriting along with manually curated benchmarks, both of which we provide open-access/source.\footnote{\url{https://github.com/anonymous-until-publication/NeuTralRewriter}} First, a rule-based rewriter is implemented leveraging hand-written rules and an automatic error correction tool. Next, a neural rewriter is trained on output generated by the rule-based rewriter to remove the need for extensive pre-processing and the reliance on computationally expensive tools such as dependency parsers. Our manual and automatic evaluation show how the neural rewriter clearly improves over the rule-based approach with word error rates (WER) below 0.18\% on synthetic, in-domain and out-domain test sets.

The main contributions of our work can be summarized as follows: (i) WinoBias+, an open-source manually curated extension of WinoBias~\citep{zhao2018gender} providing neutral alternatives for 3,167 sentences as well as a manually curated set of 1,000 natural sentences (domain: Reddit, OpenSubtitles), (ii) open-source code for rule-based and neural neutral rewriters which can convert (binary) gendered English sentences into their gender neutral counterparts, (iii) a detailed manual and automatic evaluation of errors made by the rule-based and neutral rewriter on synthetic and natural data.

\section{Related Work}\label{sec:relwork}

Recent years have seen an increase in research on gender and gender bias mitigation in NLP. While a relatively large body of research has focused on debiasing word embeddings \cite[e.g., ][]{bolukbasi2016man,Font2019,zhao2018learning}, our work is related to the generation of gender variants. We broadly distinguish between: (i) approaches that incorporate additional (meta-) information during training/testing allowing for a controlled generation of gender alternatives, and (ii) approaches that focus on gender rewriting. The synopsis will focus specifically on research related to the gender of human referents.

% Current approaches to gender bias can be divided into two main research directions: (i) bias mitigation during model training and, (ii) data augmentation by either adding meta- or counterfactual data to the training set (in a pre-processing step) or by reinflecting the monolingual output to include multiple options (in a post-processing step). We will focus on the latter as these approaches are more closely related to our work.

Within the field of Machine Translation (MT), \citet{vanmassenhove2018europarl,vanmassenhove2019getting}, and \citet{basta2020extensive} incorporate meta-information in the form of gender tags on the source side to enable gender alternative target translations for ambiguous source sentences. \citet{Moryossef2019} propose a black-box approach by appending gender information to the target sentences using parataxis constructions at translation time. \citet{Bau2019} describe work on controlling linguistic features (a.o. gender) in Neural MT by identifying and (de)activating the relevant neurons. They show that gender is the most difficult feature to control with a success rate of 21\% using the top five identified neurons.

\citet{lu2020gender} uses a Counterfactual Data Augmentation (CDA) technique to augment data sets by creating gender alternative sentences to decrease gender bias. Their approach consists of swapping gendered words with their male/female counterparts (e.g. he$:$she, father$:$mother...). Their results indicate that a CDA approach outperforms a simple word embedding debiasing technique~\cite{bolukbasi2016man}. \citet{habash2019automatic} and \citet{alhafni2020gender} present gender-aware reinflection models for Arabic. Using an Arabic sentence and a target gender, the desired gender alternative is generated by re-inflecting the input.

% \citet{zhao2018learning} and \citet{zmigrod-etal-2019-counterfactual} describe slightly different implementations of CDA techniques, the former focusing solely on swapping personal pronouns and the latter on animate nouns. All these approaches present CDA as a pre-processing technique to augment training data but could technically also be used as rewriters in a post-processing step. 

It is worth noting that all the previously described approaches focus on generating binary (female/male) gendered alternatives or translations, while our work focuses on generating gender-neutral alternatives. As such, the work that is most closely related to ours is \citet{sun2021they}. Their work is contemporaneous to our submission.\footnote{Currently in $arxiv$ pre-print.} \citet{sun2021they} present a rule-based and neural rewriter for the generation of gender-neutral \textit{singular they} sentences as well as an evaluation benchmark\footnote{We contacted the authors to obtain their benchmark for comparison as it is currently not open-source, but have not been able to obtain it yet. We will nevertheless attempt to compare our result to theirs to the best of our ability.} of 500 parallel sentences (gendered and gender-neutral) from five domains (Twitter, Reddit, movie quotes, jokes). Their rule-based and neural rewriters are able to generate gender-neutral sentences with an error-rate below 1\% (0.63\% and 0.99\% respectively). In terms of resources, compared to \citet{sun2021they}, we provide larger synthetic and natural benchmarks. In terms of performance, although complicated due to the lack of a publicly available benchmark, our models are seemingly better with error-rates of 0.52 (rule-based) and 0.02 (neural) on the most comparable benchmark, i.e. Reddit data.

% A related line of workless directly applicable to translation focuses on reinflecting monolingual sentences to reduce thelikelihood of bias (Habash et al., 2019; Alhafniet al., 2020; Sun et al., 2021).  An important dis-tinction between our work and earlier reinflectionapproaches is that we only use a single model foreach language pair, significantly reducing compu-tation and complexity.

% We will focus specifically on those approaches that are most relevant to our work. These include work on gender control in translations, gender as a correction problem and the reinflection of gendered sentences.

% In the literature on gender in NLP, two main approaches for bias mitigation can be identified: (a) approaches that attempt to mitigate bias during model or word representation training, and/or (b) approaches that aim to augment the data by creating more variety in the training set (pre-processing step) or in the output (post-processing step). In the following paragraphs, we focus on the latter as it is most closely related to our approach.
\section{Use Cases}\label{sec:usecases}
Generating neutral alternatives for gendered sentences has applications for various monolingual language generation tasks (e.g. automatic responses), where (i) one does not want to assume the gender of the referents, or (ii) one wants to present the user with various options. 
Similarly, in a bilingual setting, more specifically for MT, a neutral rewriter allows for the generation of gender neutral alternatives for genderless and gender-neutral source languages (Hungarian, Turkish, Persian, Swahili...) or null-subject source languages (Spanish, Chinese, Arabic, Bulgarian...). For illustration, Example~(\ref{ex:MT2}) and~(\ref{ex:MT})  demonstrate how gender-neutral alternatives can be useful in bilingual settings. Example~(\ref{ex:MT2}) features a sentence in Armenian using the epicene (gender-neutral) pronoun `\textit{\textarmenian{Նա}}' which can be either translated into `he', `she' or singular `they'.

\begin{liout}
\label{ex:MT2}
HY: \textarmenian{Նա բացեց դուռը}\\
EN: \textbf{He}/She/They \textbf{opened the door.}\footnote{The translation in bold is the only one provided by Bing and Google Translate consulted on May 4, 2021.} 
\end{liout}

Similarly, Example~(\ref{ex:MT}) illustrates the possible translations of a null-subject source in Spanish which can be translated as "works in a company".

\begin{liout}
\label{ex:MT}
ES: Trabaja en una empresa.\\
EN: \textbf{He}/She \textbf{works in a company.}\footnote{The translation in bold is the only one provided by Bing, Google Translate and DeepL consulted on May 4, 2021.}\\  
EN: They work in a company.        
\end{liout}

As a pre-processing step, rewriting into neutral alternatives could be useful to debias training data and thereby its embeddings \cite[see a.o.,][]{bolukbasi2016man,li2018towards,gonen2019lipstick} and/or to obfuscate sensitive `gender' features from real user data facing automatic profiling systems \cite{reddy2016obfuscating,shetty2018a4nt,emmery2021adversarial}.

\section{Methodology \& Experimental Setup}\label{sec:methodology}
% We first provide a detailed description of the training data and evaluation benchmarks (Section \ref{subsec:data}). In Section \ref{subsec:rbr}

% The Rule-Based Rewriter (RBR) approach relies on computationally expensive tools (tagger, parser), as described in Section \ref{subsec:rbr}. Therefore, we additionally experimented with a Neural Rewriter (NR)~ (Section \ref{subsec:nr}) trained on data generated by the RBR. The NR approach could remove the need for extensive pre-processing and thus provide a more scalable/generalizable rewriter.

\subsection{Datasets}\label{subsec:data}
All data is preprocessed using the Moses (de)tokenizer \cite{koehn2007moses}. Training (Reddit) and test sets (WinoBias+, OpenSubtitles, Reddit) contain a balanced amount of the eight (binary) target pronouns/determiners: \textit{he}, \textit{she}, \textit{her(s)}, \textit{his}, \textit{him}, \textit{him/herself}.\footnote{For a set containing $X$ sentences, we extracted at least $X/8$ sentences containing each form - a completely uniform distribution was not achievable due to the fact that multiple pronouns/determiners can be present in a single sentence.}

% \subsubsection{Training data}
\paragraph{Reddit} 

%A set of 71M sentences containing binary target pronouns/determiners 

A set of 2,259,386 sentences (containing a total of 3M pronouns/determiners) was randomly sampled from Pushshift's Reddit snapshots \cite[][including all subreddits]{baumgartner2020pushshift} for the period of July--December 2019. This set we would later use for training our neural rewriter. Another set of 1,693 sentences (containing a total of 2K pronouns/determiners) extracted from Reddit in the same way would later be used as a development set. There are no overlaps between the two sets.

% From this set, we randomly sampled .

% Given that there are 8 target forms, we extracted at least 375 000 (=3M/8) sentences for each form aiming to balance the distribution over the different pronouns/determiners.}

% We ought to note that sentences that contain more than one pronoun are counted multiple times -- one for each pronoun they contain; but they are included only once, that is why the amount of total extracted sentences is smaller than 3 000 000. 
% \subsubsection{Evaluation Benchmarks}
% We evaluated our approaches on the following data sets:
\paragraph{WinoBias+} an extension of the WinoBias benchmark, providing (manual) neutral alternatives for its 3,167 synthetic sentences, and corrections (e.g. for ungrammatical sentences\footnote{For example, the original WinoBias sentence ``The laborer handed the application to the editor because she \textit{want} the job.'' is corrected into ``The laborer handed the application to the editor because she \textit{wanted} the job.''}) of the original dataset.

\paragraph{OpenSubtitles, Reddit test} additional sets of 1,000 (manually corrected) parallel sentences (500 for each set). 
The entire cleaned and extended version of the corpus---\textit{WinoBias+}--- the OpenSubtitles~\cite{lison2016opensubtitles2016}, and Reddit benchmark is made publicly available under a CC BY-SA 4.0\footnote{\url{https://creativecommons.org/licenses/by-sa/4.0/}} license.\footnote{\url{https://github.com/vnmssnhv/NeuTralRewritter}}

\subsection{Rule-Based Rewriter}\label{subsec:rbr}
The rule-based rewriter (RBR), consists of two main components: (i) a rule-based pronoun rewriter, and (ii) an error-correction language model.
\subsubsection{Rule-Based Pronoun Rewriter}
Table~\ref{tbl:pronouns} gives an overview of the binary forms and their gender-neutral alternatives. While most mappings are one-to-one, `\textit{her}' can be either a pronoun (e.g. `I gave it to her.' $\rightarrow$ `I gave it to them.') or a possessive determiner (e.g. `It is her book.' $\rightarrow$ `It is their book') and `\textit{his}' can be either a possessive determiner (`It is his book.' $\rightarrow$ `It is their book') or an independent possessive pronoun (`The book is his.' $\rightarrow$ `The book is theirs'). To disambiguate these forms, the POS tagger and dependency parser from Stanza~\cite{qi2020stanza} were used.\footnote{The `his' ambiguity can only be resolved using a dependency parser since the xpos and upos tags do not differ when `his' is used as a independent or dependent possessive.}

\begin{table} % [ht]
    \small
    \centering
    {\setlength\tabcolsep{4pt}\begin{tabular}{|ccc|}\hline
    %  &  & \# of  & \# of running \\
         binary  & $\rightarrow$ & gender-neutral  \\ \hline \hline
         he, she & $\rightarrow$ &  they \\ \hline
         him  & $\rightarrow$ & them \\ \hline
         her  & $\rightarrow$ & them, their \\ \hline
         his & $\rightarrow$ & their, theirs \\ \hline
         hers & $\rightarrow$ & theirs \\ \hline
         him/herself & $\rightarrow$ & themselves\tablefootnote{`Themselves' is preferred over `themself' according the APA guidelines: \url{https://apastyle.apa.org/style-grammar-guidelines/grammar/singular-they}} \\ \hline
    \end{tabular}}
    \caption{Mapping binary pronouns/determiners to their gender-neutral alternatives.}\label{tbl:pronouns}
\end{table}

Following the guidelines from the European Parliament for gender neutral language \footnote{\url{https://www.europarl.europa.eu/cmsdata/151780/GNL_Guidelines_EN.pdf}}, we provide an option to change gendered English animate nouns (`chair(wo)man' $\rightarrow$ `chairperson', `bar(wo)man' $\rightarrow$ `bartender'...), unnecessary feminine forms of animate nouns (e.g. `actress' $\rightarrow$ `actor', `heroine' $\rightarrow$ `hero'...), and generic uses of `man' (e.g. `freshman'$\rightarrow$`first-year student', `man-made' $\rightarrow$ `human-made'...).\footnote{The complete list of nouns included can be found in the appendix.} 

% and titles (`Ms' instead or `Mrs' or `Miss') into their gender neutral counterparts.\footnote{The complete list of nouns included can be found in the appendix.}

\subsubsection{Subject-Verb Agreement Correction}

\begin{table} % [ht]
    \small
    \centering
    {\setlength\tabcolsep{4pt}\begin{tabular}{|ccc|}\hline
    %  &  & \# of  & \# of running \\
         3$^{rd}$ person  & $\rightarrow$ & plural  \\ \hline \hline
         work\textbf{s} & $\rightarrow$ &  work \\ \hline
         ha\textbf{s}  & $\rightarrow$ & have \\ \hline
         \textbf{is}  & $\rightarrow$ & are \\ \hline
    \end{tabular}}
    \caption{Subject-verb agreement correction examples.}\label{tbl:data-stat}
    \label{tbl:sv}
\end{table}

The nominative pronouns (\textit{he} and \textit{she}) can be replaced by \textit{they}. However, if they are in agreement with a simple present tense verb (or the verb `to be' ) the 3$^{rd}$ person form/ending should be replaced by a plural one (see Table~\ref{tbl:sv}). To address this, we used a Python wrapper for LanguageTool, an open-source grammar, style and spell corrector. \footnote{ \url{https://pypi.org/project/language-tool-python/}} We limited the correction to grammar mistakes to avoid additional changes (e.g. insertion of commas, different word choices, removal of whitespaces...).

\subsection{Neural Rewriter}\label{subsec:nr}
% We started from the reddit dataset of 71M sentences in English which contain personal pronouns. From this dataset we extracted 2 253 257 sentences to ensure that there are 3 000 000 pronouns included. Given that there are 8 pronouns, we extracted at least 375 000 ( = 3 000 000 / 8 ) sentences for each pronoun aiming to balance the distribution over the different pronouns. We ought to note that sentences that contain more than one pronoun are counted multiple times -- one for each pronoun they contain; but they are included only once, that is why the amount of total extracted sentences is smaller than 3 000 000. 

% The extracted sentence were designated as \emph{source} data. They were first tokenized with Moses tokenizer. 
% The set of neutral variants was designated as the \emph{target} data. We repeated this process to generate a validation set containing 1 000 pronouns; the validation set consists of 824 original-neutral sentence pairs.
We trained a Transformer model~\cite{Vaswani2017-Transformer} using \textsc{fairseq}~\cite{Ott2019-fairseq}---following the setup of ~\cite{sun2021they} for comparison. For training we used the 2,259,386 Reddit sentences as source and their gender-neutral alternatives as target; for validation we used the 1,693 Reddit sentences and their neutral alternatives (see Section~\ref{subsec:data}). The gender-neutral alternatives, i.e. the target sides, are generated by applying the RBR on the original dataset. All hyperparameters and their values are listed in the Appendix along with the preprocessing and training commands and options. %For validation, a subset of the Reddit training data was processed by the RBR to generate a parallel corpus of 1,693 sentences with gender neutral alternatives, containing 2,000 pronouns/determiners. %1,581 sentences with gender neutral alternatives, containing 1,000 pronouns/determiners.

% Using the aforementioned training and validation data we trained a transformer model~\cite{Vaswani2017-Transformer} with the fairseq~\cite{Ott2019-fairseq} framework. For comparison reasons we followed the setup of~\cite{sun2021they}. 
% In particular, we selected the \verb|transformer-iwslt-en-de| architecture with 4 attention heads and encoder and decoder embedding dimensions equal to 512 and encoder and decoder embedding dimensions for the FFN equal to 1024; we used the Adam learning optimizer~\cite{Kingma2014-Adam} with a learning rate of 0.005 and inverse square-root schedule with 4 000 warmup steps. We also employed an early stopping based on the improvement on the validation set, with patience 5. We added a dropout of 0.3. We built the vocabularies using joint (on the concatenation of the source and the target) byte-pair encoding~\cite{Sennrich2016_BPE} with 32 000 operations. We used token-based batches with maximum size of 4096. For ease of replicability we provide our complete preprocessing and training scripts in Appendix~\ref{sec:appenidx}.

\section{Results \& Discussion}\label{sec:results}
Both rewriters were (manually) evaluated on synthetic (WinoBias+) and natural (Reddit, OpenSubs) evaluation benchmarks.
\subsection{Manual Evaluation}\label{subsec:maneval}
Table~\ref{tbl:errors} presents a detailed overview of the errors per test set for the Rule-Based and Neural approach. An overview and explanation of all error labels can be found in the Appendix.
\paragraph{Rule-Based Approach}
The errors can be divided broadly into ``language model'' (LM), ``postag'' (POS) and ``other'' errors. WinoBias+ consists of 3167 sentences. Only 21 of the synthetic sentences were rewritten incorrectly. Issues arose either due to incorrect disambiguation (`her' $\rightarrow$ `them' (pronoun) instead of `their' (determiner)) or due to incorrect subject-verb agreement (SVA). 

The RBR struggled more with the noisy, often ungrammatical, natural data from OpenSubtitles and Reddit. The main issues observed are incorrect SVA, additional corrections by the language tool (unrelated to gender neutrality, e.g. \textit{cause} $\rightarrow$ \textit{because}) and incorrect disambiguation of "`s".\footnote{e.g. \textit{He's worked.} $\rightarrow$ \textit{They are worked.} instead of \textit{They have worked.}} 

\begin{table} % [ht]
    \small
    \centering
    {\setlength\tabcolsep{3pt}\begin{tabular}{|l|l|c|c|c|c|c|c|}\hline
    %  &  & \# of  & \# of running \\
         \multicolumn{2}{|c|}{\multirow{2}{*}{Error Classes}} & \multicolumn{3}{c|}{Rule-Based}   & \multicolumn{3}{c|}{Neural}   \\ \cline{3-8}
        \multicolumn{2}{|c|}{} & WB$+$ & OpenS & Red & WB$+$ & OpenS & Red       \\ \hline \hline  
         \multirow{5}{*}{\textbf{LM}}   & SVA       &9  &16 &11 &0 &5 &0   \\ \cline{2-8}  
                                        & corr.     &0  &0  &11 &0 &0 &0   \\ \cline{2-8}  
                                        & `s (has)  &0  &1  &7 &0 &1 &0   \\ \cline{2-8}
                                        & space     &0  &0  &3 &0 &0 &4   \\ \cline{2-8}
                                        & other     &0  &0  &3 &0 &0 &0   \\ \cline{2-8} \hline
         \multirow{2}{*}{\textbf{POS}}  & error     &12 &0  &3 &0 &0 &0   \\ \cline{2-8}  
                                        & source    &0  &0  &2 &0 &0 &0   \\ \cline{2-8} \hline
         \multirow{2}{*}{\textbf{OTH.}} & cap.      &0  &4  &2 &0 &1 &0   \\ \cline{2-8}
                                        & ungram.         &0  &2  &0 &0 &0 &0   \\ \cline{2-8} \hline
                                        & rule      &0  &1  &1 &0 &1 &0   \\ \cline{2-8} \hline
                                        & UNK       &0  &0  &0 &0 &0 &2   \\ \cline{2-8} \hline
         \multicolumn{2}{|c|}{\textbf{\# of errors}} &21 &24 &43 &0 &8 &6   \\ \hline   

    \end{tabular}}
    \caption{Error classification and counts on the WinoBias+, OpenSubtitles and Reddit test set for the Rule-Based and Neural approach.}\label{tbl:errors}
\end{table}
\paragraph{Neural Approach} Interestingly, and in contrast with the findings described in \citet{sun2021they}, our neural model trained on the rule-based generated training data, outperforms the rule-based approach. The error analysis reveals that the neural model resolves many of the longer distance SVA issues, the disambiguation of ``'s'' and errors that occurred due to incorrect postags.  

No errors were made on the synthetic WinoBias+ data. Errors on the in-domain Reddit data were due to the removal of additional spaces (4 errors) or because of an unknown character/emoji (2 errors). On the out-of-domain OpenSubtitles set, we noted 8 errors the majority of which due to incorrect SVA (5 errors).

\subsection{Automatic Evaluation}\label{subsec:auteval}
For comparison, we employed the same metric as \citet{sun2021they}: WER. A combination of the baseline WER (indicating the amount of changes needed in order to change to gender-neutral alternatives), and the WER computed between the correct neutral forms and the automatically generated forms provides insights into the performance of both approaches. 

% However, unlike our experiments, \citet{sun2021they} limit rewriting to sentences containing only one gendered pronoun. This most likely reduces the length of the sentences in their evaluation benchmark. 

% automatically evaluated the output of the rewriters using word error rate (WER)\citet{sun2021they}. % and BLEU \cite{Papineni2002}. 

\begin{table} % [ht]
    \small
    \centering
    {\setlength\tabcolsep{4pt}\begin{tabular}{|l|c|c|c||c|}\hline
    %  &  & \# of  & \# of running \\
        \textbf{WER (\%)}  & WB$+$ & OpenS & Reddit & \citet{sun2021they}  \\ \hline \hline
         BASE   &8.76           &14.09          &11.02          &12.40  \\ \hline 
         RBR    &0.06           &0.45  &0.52           &0.63  \\ \hline
         NR     &\textbf{0.00}  &\textbf{0.18}           &\textbf{0.02}  & 0.99 \\ \hline 
        %  Error  &0.009          & 0.037         & 0.051 \\ \hline
        %==> OpenSubsReddit.translation.wer <==
        % 0.00015629884338855892
        % ==> Reddit.translation.wer <==
        % 0.00023185717597959656
        % ==> WinoBias+.translation.wer <==
        % 0.0(
    \end{tabular}}
    \caption{WER on the synthetic WinoBias+ (WB$+$) test set and natural Reddit and OpenSubtitles benchmark vs WER obtained by \citet{sun2021they}.}\label{tbl:WER}
\end{table}

Given that \citet{sun2021they} use an evaluation benchmark of 500 sentences consisting of Twitter, Reddit, jokes and movie quotes data, its performance is probably most comparable to the scores we obtained on the Reddit set. Like the manual evaluation, and in contrast with \citet{sun2021they}, the automatic evaluation (Table \ref{tbl:WER}) confirms that our neural approach is able to generalize over the rule-based generated data, outperforming it with error rates below 0.18\% (0.0\% (WB+), 0.18\% (OpenSubtitles) and 0.02\% (Reddit)). Furthermore, these error rates are all substantially lower than those reported by \citet{sun2021they}. We hypothesize this is due to the better performance of the RBR (confirmed as well by the automatic/manual evaluation) leading to better source (gendered)--target (neutral) training data for the NMT model. 

We ought to note that WER does not take into account the removal of superfluous spaces (e.g. before the first character of a sentence, double spaces instead of a single one). We only observed the removal of such spaces by the neural rewriter on the Reddit data (see detailed manual analysis presented in Table~\ref{tbl:errors}).

\section{Conclusion}\label{sec:conclusion}
This paper presents a rule-based and a neural gender-neutral rewriter for English. First, the rule-based approach was implemented, leveraging hand-written rules and an automatic error correction tool. Using the RBR, we generated a parallel gendered-to-neutral corpus on which an NMT system was trained. The NMT model removes the need for computationally expensive pre-processing steps and, according to the manual and automatic evaluation, outperforms the RBR on synthetic, in-domain and out-domain benchmarks. Along with our open-access/source code, we also provide three manually curated benchmarks for neutral rewriting.

For now, the neutral rewriter is limited to English using `singular they' and recommendations for gender neutral writing specific to the English language. It is, in theory, possible to extend this approach (or a similar one) to other languages. However, so far, few languages have a crystallized approach when it comes to gender-neutral pronouns and gender-neutral word endings. 

In future work, we intend to explore potential applications of the neutral rewriters (e.g. gender debiasing of corpora). We furthermore plan to extend our work to gender-neutral rewriting targeting specific referents within a sentence to accommodate the gender preferences of individual referents. 

\section*{Ethics statement}
\paragraph{Neutral Rewriter Application} The Neutral Rewriter is intended to provide gender-neutral alternatives and increase the inclusiveness of NLP/MT applications. The rewriter can furthermore be used as a preprocessing step to obfuscate a potentially sensitive gender attribute from training data. 

At this stage, the rewriter works on a sentence-level and does not allow for rewriting pronouns or determiners of specific referents. We followed the guidelines of the European Parliament for gender neutral language and provide an option to change gendered animate nouns, unnecessary feminine forms of animate nouns and generic uses of the word `man' based on non-exhaustive word lists.

\paragraph{Datasets} We present three openly available English benchmarks: (i) WinoBias+, (ii) OpenSubtitles and (iii) Reddit. (i) WinoBias+ consists of a curated and extended version of the synthetic WinoBias~\cite{zhao-etal-2018-gender} dataset, distributed under the MIT License.\footnote{\url{https://opensource.org/licenses/MIT}}
(ii) The open-source OpenSubtitles~\cite{lison2016opensubtitles2016}\footnote{\url{http://www.opensubtitles.org/}} data was used to randomly sample a subset for the OpenSubtitles benchmark. OpenSubtitles is distributed under a Creative Commons license.\footnote{Attribution-Non Commercial 4.0 International} (iii) The Reddit dataset was collected through the third-party snapshots of Reddit's publicly available API at \url{https://pushshift.io}. It is subject to Reddit's own User Agreement and Privacy Policy and covers the \emph{free and public sharing of user data}.\footnote{See \url{https://www.redditinc.com/policies/user-agreement} and \url{https://www.redditinc.com/policies/privacy-policy} respectively.}

The neutral alternatives for the three benchmarks were manually created by a linguist. The curation rationale behind the selected datasets is summarized as follows. WinoBias was selected as it is one of the few benchmarks for gender bias in NLP. We extended it with gender-neutral alternatives. The natural Reddit and OpenSubtitles dataset allowed us to verify the robustness of the rewriters on more noisy and diverse data sets. The OpenSubtitles and Reddit datasets contain variety in terms of language and English social dialects. Training and test sets contain a balanced amount of the eight (binary) target pronouns/determiners. For a set containing $X$ sentences, we extracted at least $X/8$ sentences containing each form - a completely uniform distribution was not achievable due to the fact that multiple pronouns/determiners can be present in a single sentence.

\paragraph{Carbon statement}
The neural model presented in this work has an ecological footprint equivalent to 1.68kg of CO2 emissions.\footnote{Contribution based on GPU power consumption at training the NeuTral rewriter model.} The training time, consumption and carbon emission can be found in Table~\ref{tbl:carbon}.

% Describe the characteristics of the dataset in enough detail for a reader to understand which speaker populations the technology could be expected to work for. (For suggestions of what kind of information to include, see Bender and Friedman 2018, Mitchell et al 2019 and Gebru et al 2018.)

% INFO ETHICAL STATEMENT 
% Ensure that your dataset has been collected in a manner which is consistent with the terms of use of any sources and the intellectual property and privacy rights of the original authors of the texts. The fact that a dataset has already been used previously does not necessarily make it acceptable. If appropriate, check that informed consent was signed. Your institution may have a data officer or a review board charged with helping researchers navigate these issues and we encourage you to reach out to them for assistance. Such reviews may take time, so we advise you to contact them early if needed.

\begin{table}[]
    \centering
    \small 
    \begin{tabular}{|c|c|c|c|c|}\hline
 Elapsed & Avg. power & kWh & CO2\\
 time (h) & draw & &  (kg) \\\hline
 6.2 & 147.37 & 2.64 & 1.68 $\pm$0.13\\\hline
    \end{tabular}
    \caption{Train time, consumption and carbon emissions related to the training of the NeuTral rewriter.}
    \label{tbl:carbon}
\end{table}

\section*{Acknowledgements}
We would like to thank the reviewers for their insightful feedback and comments.

\bibliographystyle{acl_natbib}
\bibliography{custom}
\appendix
% This must be in the first 5 lines to tell arXiv to use pdfLaTeX, which is strongly recommended.
\pdfoutput=1
% In particular, the hyperref package requires pdfLaTeX in order to break URLs across lines.

% \documentclass[11pt]{article}

% \usepackage[hyperref]{emnlp2021}
% % Standard package includes
% \usepackage{times}
% \usepackage{latexsym}
% \usepackage{arydshln} 
% \usepackage{graphicx}
% \usepackage{multirow}
% \usepackage{subfigure}
% \usepackage{supertabular}

% \renewcommand{\UrlFont}{\ttfamily\small}

% \newcounter{example}[section]
% \newenvironment{example}[1][]{\refstepcounter{example}\par\medskip
%   \noindent \textbf{Example~\theexample. #1} \rmfamily}{\medskip}

% For proper rendering and hyphenation of words containing Latin characters (including in bib files)
% \usepackage[T1]{fontenc}
% % For Vietnamese characters
% % \usepackage[T5]{fontenc}
% % See https://www.latex-project.org/help/documentation/encguide.pdf for other character sets

% % This assumes your files are encoded as UTF8
% \usepackage[utf8]{inputenc}

% % This is not strictly necessary, and may be commented out,
% % but it will improve the layout of the manuscript,
% % and will typically save some space.
% \usepackage{microtype}

% % If the title and author information does not fit in the area allocated, uncomment the following
% %
% %\setlength\titlebox{<dim>}
% %
% % and set <dim> to something 5cm or larger.

% % \title{Appendix}

% % \date{}

% \begin{document}
% \maketitle

\twocolumn

\section{Appendix}\label{sec:appendix}
The appendix provides additional information on generation of gender-neutral alternatives (Section~\ref{subsec:advrewriter}), the error labels and analysis (Section~\ref{subsec:erroranalysis}) and the training hyperparameters of the Neural Machine Translation model (Section~\ref{subsubsec:hyperparams}).

\subsection{Advanced Rewriter}\label{subsec:advrewriter}
The advanced rewriter includes rewriting of gender-marked job titles (chairman, anchorman...), rewriting of unnecessary feminine forms (actress, comedienne, waitress...), avoidance of construction using a generic form of `man' (`average man', 'man and wife'...), and rewriting of titles (`Mrs' and `Miss').

\subsubsection{Gender-neutral alternatives for gender-marked job titles}
\begin{table}[ht]
    \small
    \centering
    \resizebox{\columnwidth}{!}
    {\setlength\tabcolsep{2pt}\begin{tabular}{|lcl||lcl|}\hline
    %  &  & \# of  & \# of running \\
        \multicolumn{3}{|c|}{chairman/woman}                   &    \multicolumn{3}{c|}{businessman/woman}                           \\ \hline \hline
        chairman        & $\rightarrow$     & chairperson      &    businessman     & $\rightarrow$     & business person           \\ \hline 
        chairmen        & $\rightarrow$     & chairpeople      &    businessmen     & $\rightarrow$     & business people           \\ \hline 
        chairwoman      & $\rightarrow$     & chairperson      &    businesswoman   & $\rightarrow$     & business person           \\ \hline 
        chairwomen      & $\rightarrow$     & chairpeople      &    businesswomen   & $\rightarrow$     & business people           \\ \hline \hline
        \multicolumn{3}{|c|}{anchorman/woman}                    &   \multicolumn{3}{c|}{postman/postwoman}                           \\ \hline \hline
        anchorman       & $\rightarrow$     & anchor           &    postman         & $\rightarrow$     & mail carrier              \\ \hline 
        anchormen       & $\rightarrow$     & anchors          &    postmen         & $\rightarrow$     & mail carriers             \\ \hline 
        anchorwoman     & $\rightarrow$     & anchor           &    postwoman       & $\rightarrow$     & mail carrier              \\ \hline 
        anchorwomen     & $\rightarrow$     & anchors          &    postwomen       & $\rightarrow$     & mail carriers             \\ \hline \hline
        \multicolumn{3}{|c|}{congresswoman/congressman}          &    \multicolumn{3}{c|}{mailman/mailwoman}                           \\ \hline \hline
        congressman     & $\rightarrow$     & member of congress  & mailman         & $\rightarrow$     & mail carrier              \\ \hline 
        congressmen     & $\rightarrow$     & members of congress & mailmen         & $\rightarrow$     & mail carriers             \\ \hline 
        congresswoman   & $\rightarrow$     & member of congress  & mailwoman       & $\rightarrow$     & mail carrier              \\ \hline 
        congresswomen   & $\rightarrow$     & members of congress & mailwomen       & $\rightarrow$     & mail carriers             \\ \hline \hline
        \multicolumn{3}{|c|}{policeman/policewoman}               &   \multicolumn{3}{c|}{salesman/saleswoman}                         \\ \hline \hline
        policeman       & $\rightarrow$     & police officer    &   salesman        & $\rightarrow$     & salesperson               \\ \hline 
        policemen       & $\rightarrow$     & police officers   &   salesmen        & $\rightarrow$     & salespersons              \\ \hline 
        policewoman     & $\rightarrow$     & police officer    &   saleswoman      & $\rightarrow$     & salesperson               \\ \hline 
        policewomen     & $\rightarrow$     & police officers   &   saleswomen      & $\rightarrow$     & salespersons              \\ \hline \hline
        \multicolumn{3}{|c|}{spokesman/woman}                    &    \multicolumn{3}{c|}{fireman/firewoman}                           \\ \hline \hline
        spokesman       & $\rightarrow$     & spokesperson     &    fireman         & $\rightarrow$     & firefighter               \\ \hline 
        spokesmen       & $\rightarrow$     & spokespersons    &    firemen         & $\rightarrow$     & firefighters              \\ \hline 
        spokeswoman     & $\rightarrow$     & spokesperson     &    firewoman       & $\rightarrow$     & firefighter               \\ \hline 
        spokeswomen     & $\rightarrow$     & spokespersons    &    firewomen       & $\rightarrow$     & firefighters              \\ \hline \hline
        \multicolumn{3}{|c|}{steward/stewardess}               &       \multicolumn{3}{c|}{barman/barwoman}                             \\ \hline \hline
        steward         & $\rightarrow$     & flight attendant   &  barman          & $\rightarrow$     & bartender                 \\ \hline 
        stewards        & $\rightarrow$     & flight attendants  &  barmen          & $\rightarrow$     & bartenders                \\ \hline 
        stewardess      & $\rightarrow$     & flight attendant   &  barwoman        & $\rightarrow$     & bartender                 \\ \hline 
        stewardesses    & $\rightarrow$     & flight attendants  &  barwomen        & $\rightarrow$     & bartenders                \\ \hline \hline
        \multicolumn{3}{|c|}{headmaster/mistress}                  &  \multicolumn{3}{c|}{cleaning man/lady}                           \\ \hline \hline
        headmaster      & $\rightarrow$     & principal          &  cleaning man    & $\rightarrow$     & cleaner                   \\ \hline 
        headmasters     & $\rightarrow$     & principals         &  cleaning lady   & $\rightarrow$     & cleaners                  \\ \hline 
        headmistress    & $\rightarrow$     & principal          &  cleaning men    & $\rightarrow$     & cleaner                   \\ \hline 
        headmistresses  & $\rightarrow$     & principals         &  cleaning ladies & $\rightarrow$     & cleaners                  \\ \hline \hline
        \multicolumn{3}{|c|}{foreman/forewoman}                  &  \multicolumn{3}{c|}{}                   \\ \hline \hline
        foreman         & $\rightarrow$     & supervisor          &         &       &                       \\ \hline 
        foremen         & $\rightarrow$     & supervisors         &         &       &                       \\ \hline 
        forewoman       & $\rightarrow$     & supervisor          &         &       &                       \\ \hline 
        forewomen       & $\rightarrow$     & supervisors         &         &       &                       \\ \hline \hline
    \end{tabular}}
    \caption{Gender-neutral alternatives for gender-marked job titles}\label{tbl:genderneutral}
\end{table}

\newpage
\subsubsection{Gender-neutral alternatives for unnecessary feminine forms}
\begin{table}[ht]
    \small
    \centering
    \resizebox{\columnwidth}{!}
    {\setlength\tabcolsep{2pt}\begin{tabular}{|lcl||lcl|}\hline
    %  &  & \# of  & \# of running \\
        \multicolumn{3}{|c|}{actress}                           &    \multicolumn{3}{c|}{usherette}                           \\ \hline \hline
        actress         & $\rightarrow$     & actor             &    usherette      & $\rightarrow$     & usher           \\ \hline 
        actresses       & $\rightarrow$     & actors            &    usherettes     & $\rightarrow$     & usher          \\ \hline \hline
        \multicolumn{3}{|c|}{heroine}                           &   \multicolumn{3}{c|}{authoress}                           \\ \hline \hline
        heroine         & $\rightarrow$     & hero              &    authoress         & $\rightarrow$     & author              \\ \hline 
        heroine         & $\rightarrow$     & heroes            &    authoresses       & $\rightarrow$     & authors            \\ \hline \hline
        \multicolumn{3}{|c|}{comedienne}                        & \multicolumn{3}{c|}{mailman/mailwoman}                           \\ \hline \hline
        comedienne     & $\rightarrow$     & comedian           & mailman         & $\rightarrow$     & mail carrier              \\ \hline 
        comediennes   & $\rightarrow$     & comedians           & mailwomen       & $\rightarrow$     & mail carriers             \\ \hline \hline
        \multicolumn{3}{|c|}{executrix}                         &   \multicolumn{3}{c|}{boss lady}                         \\ \hline \hline
        executrix       & $\rightarrow$     & executor          &   boss lady        & $\rightarrow$     & boss               \\ \hline 
        executrices     & $\rightarrow$     & executors         &   boss ladies     & $\rightarrow$     & boss               \\ \hline 
        executrixes     & $\rightarrow$     & executor          &        &      &               \\ \hline \hline
        \multicolumn{3}{|c|}{poetess}                           &    \multicolumn{3}{c|}{waitress}                           \\ \hline \hline
        poetess         & $\rightarrow$     & poet              &    waitress         & $\rightarrow$     & waiter               \\ \hline 
        poetesses       & $\rightarrow$     & poets             &    waitresses       & $\rightarrow$     & waiters              \\ \hline \hline
    \end{tabular}}
    \caption{Gender-neutral alternatives for unnecessary feminine forms}\label{tbl:unnecessaryfeminine}
\end{table}

\subsubsection{Gender-neutral alternatives for generic `man'}
\begin{table}[ht]
    \small
    \centering
    \resizebox{\columnwidth}{!}
    {\setlength\tabcolsep{2pt}\begin{tabular}{|lcl||lcl|}\hline
    %  &  & \# of  & \# of running \\
        \multicolumn{3}{|c|}{average man}                                   &    \multicolumn{3}{c|}{layman}                           \\ \hline \hline
        average man             & $\rightarrow$     & average person        &    layman     & $\rightarrow$     & layperson           \\ \hline 
        average men             & $\rightarrow$     & average people        &    laymen     & $\rightarrow$     & laypeople          \\ \hline \hline
        \multicolumn{3}{|c|}{best man for the job}                          &   \multicolumn{3}{c|}{freshman}                           \\ \hline \hline
        best man for the job    & $\rightarrow$    & best person for the job &  freshman    & $\rightarrow$ & first-year student             \\ \hline 
        best men for the job    & $\rightarrow$    & best people for the job &  freshmen    & $\rightarrow$ & first-year students                         \\ \hline \hline
        \multicolumn{3}{|c|}{mankind}                        & \multicolumn{3}{c|}{man-made}                           \\ \hline \hline
        mankind     & $\rightarrow$     & humankind           & man-made         & $\rightarrow$     & human-made              \\ \hline 
        \multicolumn{3}{|c|}{workmanlike}                         &   \multicolumn{3}{c|}{man and wife}                         \\ \hline \hline
        workmanline       & $\rightarrow$     & skillful          &   man and wife        & $\rightarrow$     & husband and wife               \\ \hline 

    \end{tabular}}
    \caption{Gender-neutral alternatives for generic `man'}\label{tbl:generic}
\end{table}

\subsection{Overview Error Analysis}\label{subsec:erroranalysis}
\subsubsection{Error Classification Rewriter}
As explained in the paper, errors are divided into Language Model (LM) errors, postag error (POS) and other errors (OTHER). Within these three error classes, we identified multiple subclasses of LM, POS and OTHER errors. An explanation of the labels used in our error analysis and paper can be found in Table \ref{tab:labels}. Table \ref{tbl:exampleserrors} provides example input and output sentences.

\begin{table}[ht]
    \centering
    {\small
    \begin{tabular}{|p{0.3\linewidth} | p{0.6\linewidth}|} \hline
    \textbf{Error Label}  & \textbf{Explanation} \\ \hline
    LM 's & Wrongly disambiguated the contracted form `s as a verb form of `to be' instead of `to have'  \\ \hline
    LM space & Space added or removed by rewriter \\ \hline
    LM correction (corr.) & Error correction done by rewriter (language tool) that is not related to gender-neutral rewriting \\ \hline
    LM subject-verb agreement (SVA) & Failure to make correct subject-verb agreement, usually due to long distance dependencies. \\ \hline
    POS & Wrong form of `they' produced by rewriter due to incorrect postag \\ \hline
    POS (source) & Wrong form of `they' produced by rewriter due to incorrect postag which is related to an ungrammatical/incorrect soure sentences\\ \hline
    OTHER rule & Some forms such as `hisn's' are not standard language and does not covered by our rules. Similarly written forms such as `hes' for `he's' are not corrected by the rewriter\\ \hline
    Other ungram. & Ungrammatical input sentence leading to an ungrammatical output\\ \hline
    Other UNK & The Neural Rewriter outputs <UNK> for unknown characters (in our case "?", "!", "...", and emojis/special characters that did not appear in our Reddit training data)\\ \hline
    \end{tabular}}
    \caption{Error label explanation}
    \label{tab:labels}
\end{table}

\begin{table}[ht]
    \small
    \centering
    \resizebox{\columnwidth}{!}
    {\setlength\tabcolsep{2pt}\begin{tabular}{|l||p{0.3\linewidth} |c|p{0.3\linewidth} |}\hline
    %  &  & \# of  & \# of running \\
    \textbf{Error Label} & \textbf{Example} & \textbf{$\rightarrow$} & \textbf{Output RBR} \\ \hline
    LM (`s)          & He\textbf{`s} worked hard & $\rightarrow$ & They \textbf{are} worked hard. \\ \hline
    LM (space)       & ... aren 't...            & $\rightarrow$ & ...aren't... \\ \hline
    LM (corr.)  & Bit pricey...            & $\rightarrow$ & \textbf{A} bit pricey... \\ \hline
    LM (SVA)         & He works and works...            & $\rightarrow$ & They work and \textbf{works}... \\ \hline
    POS         & He saw her run fast...            & $\rightarrow$ & They saw \textbf{their} run fast... \\ \hline
    POS (source) & ...looked at \textbf{her weird} (she `s close.. &$\rightarrow$ & ... looked at \textbf{their weird} (they are close... \\ \hline
    Basic rule & She's \textbf{hisn`s}.. &$\rightarrow$ & .They are \textbf{hisn's} \\ \hline
    Other & Where's herself. &$\rightarrow$ & Where's themselves. \\ \hline
    \end{tabular}}
    \caption{Examples Error Labels}\label{tbl:examples}
    \label{tbl:exampleserrors}
\end{table}
\newpage

\subsection{Neural Rewriter}\label{subsec:NR}
Our neural model is trained with the following options: \verb|transformer-iwslt-en-de| architecture with 4 attention heads and encoder and decoder embedding dimensions equal to 512, encoder and decoder embedding dimensions for the FFN equal to 1024, Adam learning optimizer~\cite{Kingma2014-Adam} with a learning rate of 0.005 and inverse square-root schedule with 4 000 warmup steps, an early stopping based on the improvement on the validation set with patience 5, dropout of 0.3, joint byte-pair encoding ~\cite{Sennrich2016_BPE} with 32 000 operations, token-based batches with maximum size of 4096. For ease of replicability we provide our complete preprocessing and training scripts in Appendix.

\subsubsection{Training Hyperparameters}\label{subsubsec:hyperparams}
{\small 
\begin{verbatim}
fairseq-preprocess --source-lang $SRC \ 
  --target-lang $TRG \
  --trainpref $ENGDIR/data/train.tc.bpe \ 
  --validpref $ENGDIR/data/dev.tc.bpe \
  --testpref $ENGDIR/data/test.tc.bpe \
  --destdir $ENGDIR/data/ready_to_train
\end{verbatim}

\begin{verbatim}
fairseq-train $ENGDIR/data/train_data \
  --arch transformer_iwslt_de_en \
  --lr 0.0005 --optimizer adam \
  --adam-betas '(0.9, 0.98)' \
  --max-tokens 4096 \
  --dropout 0.3 \
  --update-freq=1 \
  --lr-scheduler inverse_sqrt \
  --warmup-init-lr 1e-07 --min-lr 1e-09 \
  --warmup-updates 4000 \
  --save-dir $ENGDIR/model \ 
  --skip-invalid-size-inputs-valid-test \
  --patience 5
\end{verbatim}
}

\noindent
With \verb|$ENGDIR| we indicate the path where the data folder and the model folder are located.

\end{document}